\documentclass[runningheads]{llncs}
\usepackage{graphicx}
\usepackage{multirow}
\usepackage{tabularx}
\usepackage{booktabs}
\usepackage{amsmath}
\usepackage{bbm}
\usepackage{adjustbox}
\newcolumntype{C}[1]{>{\centering\let\newline\\\arraybackslash\hspace{0pt}}m{#1}}
\newcolumntype{L}[1]{>{\raggedright\let\newline\\\arraybackslash\hspace{0pt}}m{#1}}
\begin{document}
\title{BackMix: Mitigating Shortcut Learning in Echocardiography with Minimal Supervision}
\titlerunning{BackMix}
%
\author{Kit M. Bransby\inst{1,2}, Arian Beqiri\inst{1}, Woo-Jin Cho Kim\inst{1}, Jorge Oliviera\inst{1}, Agisilaos Chartsias\inst{1}, Alberto Gomez\inst{1}}


%
\authorrunning{Bransby et al.}

\institute{
Ultromics Ltd., Oxford, United Kingdom
\and
Queen Mary University of London, United Kingdom
}

\maketitle              
\begin{abstract}
Neural networks can learn spurious correlations that lead to the correct prediction in a validation set, but generalise poorly because the predictions are right for the wrong reason. This undesired learning of naive shortcuts (Clever Hans effect) can happen for example in echocardiogram view classification when background cues (e.g. metadata) are biased towards a class and the model learns to focus on those background features instead of on the image content. We propose a simple, yet effective random background augmentation method called BackMix, which samples random backgrounds from other examples in the training set. By enforcing the background to be uncorrelated with the outcome, the model learns to focus on the data within the ultrasound sector and becomes invariant to the regions outside this. We extend our method in a semi-supervised setting, finding that the positive effects of BackMix are maintained with as few as 5\% of segmentation labels. A loss weighting mechanism, wBackMix, is also proposed to increase the contribution of the augmented examples. We validate our method on both in-distribution and out-of-distribution datasets, demonstrating significant improvements in classification accuracy, region focus and generalisability. Our source code is available at: https://github.com/kitbransby/BackMix

\end{abstract}
\section{Introduction}

\begin{figure}[t!]
    \centering
    \includegraphics[width=1\linewidth]{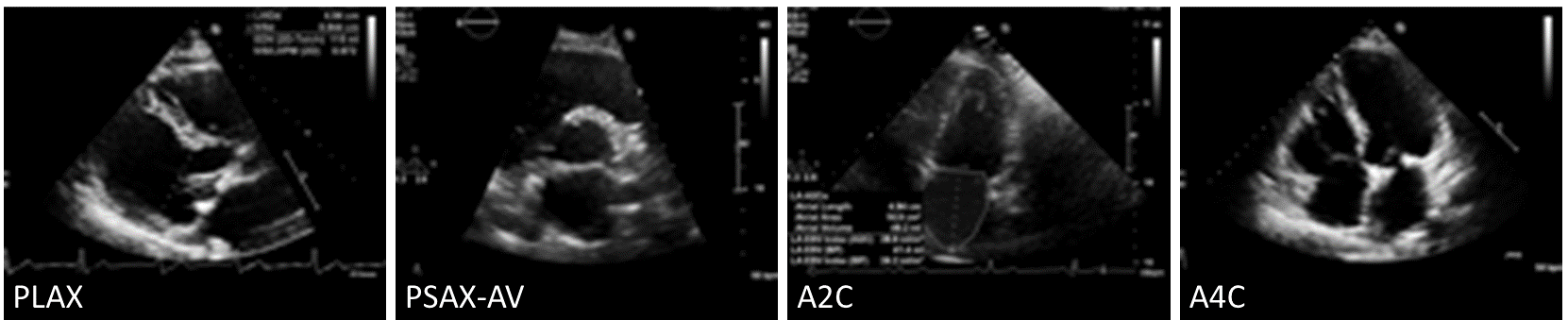}
    \caption{Examples of echocardiograms: Unwanted text and measurement data can be seen in regions outside the ultrasound sector. View label superimposed on bottom left.}
    \label{fig:class_examples}
\end{figure}

\begin{figure}[t!]
    \centering
    \includegraphics[width=1\linewidth]{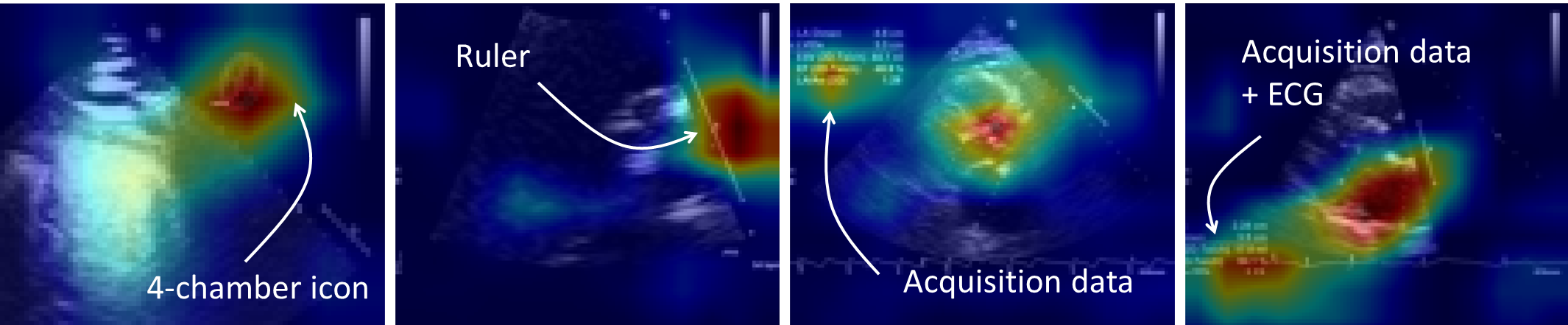}
    \caption{Examples of shortcut learning in TMED dataset induced by metadata, which can be specific to the manufacturer, software and operator, but not the patient.}
    \label{fig:tmed_gradcam}
\end{figure}

\indent Echocardiography (echo) is one of the primary cardiovascular imaging modalities used to study the structures and function of the heart from a variety of cross-sectional views (Figure~\ref{fig:class_examples}). Classification of the view is a necessary first step in automated echo analysis as views are not labelled during acquisition, and are required for reliable interpretation~\cite{vaseli2019designing}. Furthermore, the development of an accurate automated classifier is challenging due to the extensive time required to manually label studies for training data, which are often several thousand frames in total. Several convolutional neural network based methods have been used to create echo view classifiers \cite{vaseli2019designing,wegner2022accuracy,kusunose2020clinically,madani2018fast} demonstrating excellent performance. \\
\indent An echo video consists of ultrasound images framed within a triangle or trapezoid known as the ultrasound ``sector", set against a black background with patient and acquisition metadata overlaid on top. Metadata outside of the sector can spuriously correlate with the ultrasound view classification label, leading neural networks to focus upon these features instead, as visualised in the GradCAM~\cite{selvaraju2017grad} feature attribution heatmaps of Figure~\ref{fig:tmed_gradcam}. Such shortcuts allow learning of simple decision rules, thus limiting the classifier's capacity to build accurate and trustworthy heart representations. This behaviour can be difficult to detect as test data used for model validation are typically drawn from the same distribution (i.d) as training data, and can therefore leverage shortcuts leading to artificially high performance. Once deployed in the wild however, the quality of the view classifier may deteriorate when it encounters out of distribution (o.o.d) images from different medical sites, acquisition protocols or scanner manufacturers, all of which are non-patient specific and can affect metadata. \\
\indent A standard preprocessing step is to remove the area outside the sector using image segmentation, as for instance demonstrated in the CAMUS dataset~\cite{leclerc2019deep}. However, this is limited by unreliable performance in edge cases, may require training additional networks, and adds significant computation time during inference. Another perspective involves leveraging attention, which implicitly learns to focus on image regions~\cite{guo2022attention}. However without direct supervision, attention networks may also focus on spurious correlations to minimise the objective function during training. Ma et al.~\cite{ma2023rectify} address this in a natural image setting by supervising with saliency maps learnt from eye-gaze data. Despite excellent performance, this model requires an encoder-decoder as part of a multi-model pipeline, which adds to computational expense.  Bassi et al.~\cite{bassi2024improving} integrate an interpretable layer-wise relevance propagation (LRP)~\cite{bach2015pixel} module, which estimates the contribution of each pixel by using back-propagated gradients. This approach does not change the architecture and adds minimal computation; however LRP attention maps are not perfect representations, and can be noisy and non-discriminative~\cite{jung2021explaining}. \\
\indent We propose a simple yet effective random background augmentation method called BackMix, which encourages a classification network to focus on the area inside the ultrasound sector. We initially split a training image into sector and background regions, and then randomly replace the background with a background from another training example. By making the background uncorrelated to the outcome, the model learns to ignore the background and becomes invariant to the spurious regions. Our method has the advantage of not adding any parameters or architectural changes, nor does it incur any additional training or inference time. The closest works to ours have applied background blurring~\cite{li2022semantic}, foreground in-painting for segmentation~\cite{hasan2021segmentation} and image synthesis~\cite{zhong2023lightweight} in natural image settings, however ours explores the impact on o.o.d performance and is the first to be applied to non-natural medical images. \\
\indent Similarly to the above learning-based methods, training requires segmentation masks that separate ultrasound sectors from backgrounds prior to applying any augmentation. To avoid the need of acquiring segmentation masks for all training data, we extend our method to a semi-supervised setting where BackMix is only applied to a fraction of the training data. The positive effects of BackMix are maintained when as little as 5\% of the training dataset is used in augmentation. Our methodology is strengthened by re-weighting the classification loss at an example level, so that 
a higher loss is assigned to examples that use BackMix. Despite such minimal supervision we achieve a significant improvement in performance on an o.o.d dataset against a baseline classifier trained without BackMix. We evaluate our hypothesis through GradCAM~\cite{selvaraju2017grad} analysis to show both quantitatively and qualitatively that our model focuses more on the ultrasound data within the sector and ignores spurious features. \\
\indent \emph{Contributions}: We identify that shortcut learning of background metadata harms generalisability in echocardiogram view classification and propose an effective background mixing augmentation called BackMix. We explore a semi-supervised setting and demonstrate that minimal numbers of segmentation masks are required for significant improvements in classification and focus metrics. Our method is strengthened with wBackMix, which emphasises examples with random backgrounds by appropriately re-distributing the loss. We show that our method removes the need for background removal in inference, a common and computationally expensive requirement. Finally, we propose two metrics to quantitatively evaluate how much the ultrasound sector affects the prediction label.

\section{Methods}

\begin{figure}[t!]
    \centering
    \includegraphics[width=1\linewidth]{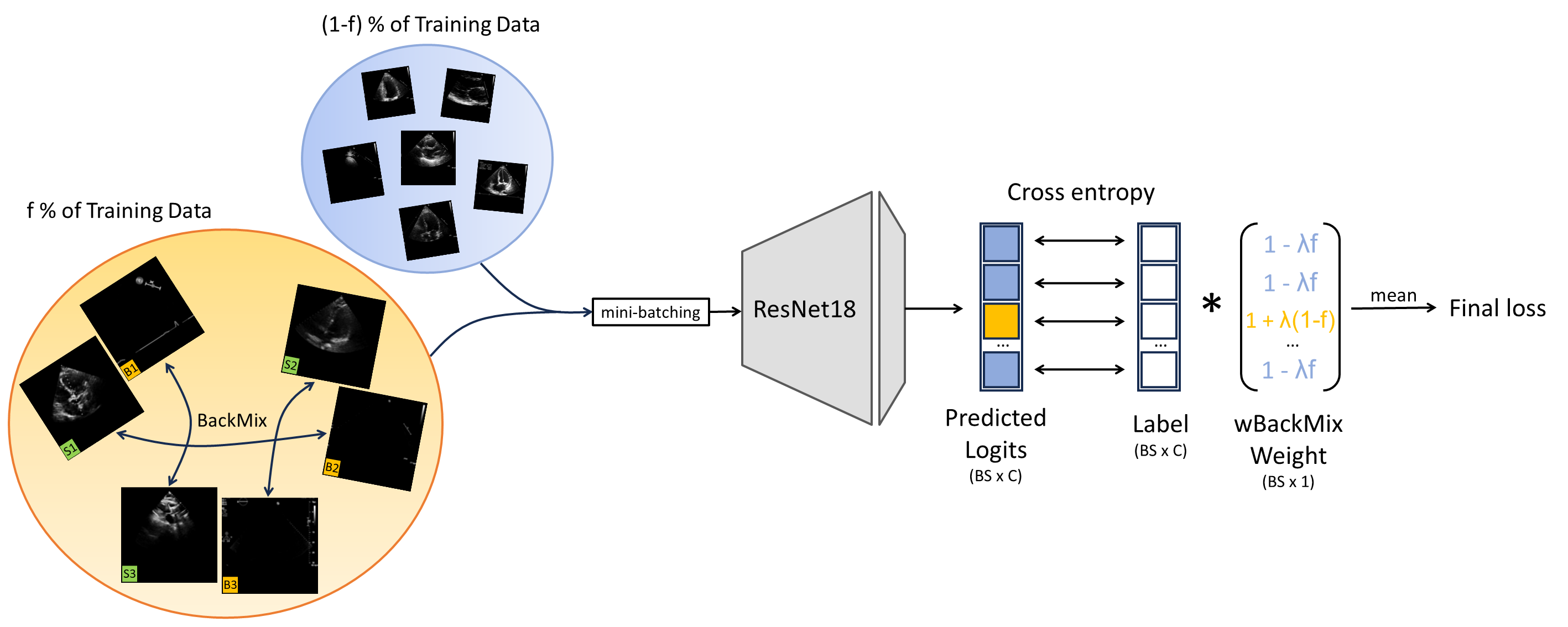}
    \caption{Training schematic: Backgrounds are shuffled between a subset of training examples, and the prediction loss re-weighted in favour of these examples}
    \label{fig:schematic}
\end{figure}

\subsection{BackMix}

We implement an augmentation method, tailored to ultrasound data, called BackMix, which randomly swaps backgrounds between images in the training set. During training, we separate an image $i$ into sector $S_{i}$ and background $B_{i}$ using a segmentation mask $M_{i}$ and in-paint the empty areas with zeros (value of background). An additional frame $j$ is then randomly sampled and the same process is applied to yield $S_{j}$ and $B_{j}$. $S_{i}$ is superimposed onto $B_{j}$ to replace $S_{i}B_{i}$ and synthesise a new sample $S_{i}B_{j}$ that is used for training. This augmentation process is illustrated in Figure~\ref{fig:schematic}. 

We apply BackMix augmentation after standard augmentations, such as random image rotations in the range [-30$^{\circ}$, +30$^{\circ}$], brightness-contrast adjustment, and horizontal flipping. BackMix is evaluated when training a ResNet18~\cite{he2016deep} network, that is selected due to its widespread use and computational efficiency, although there is no restriction in the choice of backbone network architecture.
During inference there is no need for segmentation masks. As demonstrated in Section~\ref{sec:experiments}, training with BackMix augmentations encourages the classification network to focus in the ultrasound sector.

\subsection{Semi-Supervised classification}\label{semi-supervised}

Pixel-wise segmentation labels of the ultrasound sector are time consuming and difficult to obtain as the sector boundaries are not always well defined. We thus explore a semi-supervised approach, where only a fraction $f$ of the training dataset has sector segmentation masks available. BackMix augmentation is performed only on the random $f\%$ sample of the training data, leaving the remaining $(1-f)\%$ of images untouched.\footnote{Note that during training, random backgrounds can only be sampled from frames within the $f\%$ subset and not outside.} 
The network's focus is expected to correlate with $f$ and the backgrounds pool size that participate in the augmentation. 
We observe this trend empirically, but find evidence that even with few segmentation masks, significant improvement in performance is seen when compared to networks where BackMix is disabled. 

\subsection{wBackMix}

In scenarios where $f$ is small and BackMix is only applied to few examples, the supervisory signal may be weak and overshadowed by the large quantity of examples with no BackMix. We address this by re-weighting the cross-entropy loss on an example-level to increase the contribution of augmented examples. Specifically, we devise a weighting, which scales non-augmented examples by a factor of $1 - \lambda f$ and augmented examples by $1 + \lambda (1 - f)$. Parameter $\lambda$ is a constant that is found empirically. When $\lambda$ is set to a value of zero, it weighs all examples equally, but when increased, it puts more weight on augmented examples. This weight formulation maintains the loss magnitude as the sum of the loss weight in a batch is equal to 1, mitigating any undesirable changes in the training dynamics of the model between experiments.  

\subsection{Evaluating Focus}

To determine whether each model is attending to the pixels inside the sector when making a prediction, we devise two attention-based metrics: energy percentage $\%E$, and focus percentage $\%F$. Firstly, GradCAM class activation maps are calculated for every test image $i \in I$, which give an importance score $z_{p} \in [0,1]$ to pixels $p \in i$ based on back-propagated gradients. For $\%E$, the corresponding sector mask $m \in M$ is used to compare the $z_{p}$ values of pixels inside sector $m \odot i$ with the $z_{p}$ values across the whole image $i$. This is formalised as follows, where $N$ is the test set size:

\begin{equation*}
\%E = \frac{1}{N}\sum_{i \in I, m \in M} \frac{\sum_{p \in m \odot i} z_{p}}{\sum_{p \in i}z_{p}}
\end{equation*}

Higher values of $\%E$ indicate that the network is attending more to pixels within the sector region. We also quantify the main regions of focus using $\%F$, which only considers highly activated pixels where $z_{p} > 0.5$. 
We determine the fraction of these pixels that intersects with the sector mask as follows: 

\begin{align*}
z_h = \left\{\begin{matrix} 1 & \text{if}\: \: z_{p} > 0.5 \\ 0 & \text{otherwise} \end{matrix}\right.
&&
\%F = \frac{1}{N} \sum_{i \in I, m \in M} \frac{\sum_{p \in m \odot z_h} p}{\sum_{p \in z_h} p}
\end{align*}

\section{Experiments and Results} \label{sec:experiments}

\subsection{Datasets}

We train a view classifier on TMED~\cite{huang2022tmed,huang2021new} public dataset, a collection of echo studies acquired in the course of routine care from 2011--2020 at Tufts Medical Center, Boston, USA. A labelled subset of 24,964 frames from 1,266 patients was extracted. We extensively validate the classifier generalisability on TMED test set (in-distribution dataset), and WASE Normals~\cite{asch2019need}, a large multi-site proprietary dataset (out-of-distribution dataset). WASE Normals contains 36,029 echo videos from 2,009 healthy volunteers acquired at 18 sites from 15 countries. 

Both datasets were filtered to retain the shared view labels, PLAX, PSAX-AV, A2C and A4C, and split into train (80\%), validation (10\%) and test (10\%) sets at a patient level. A single random frame was sampled from each WASE Normals video, resulting in a final dataset of 14,569 train (TMED), 1,670 validation (TMED), 1,815 i.d test (TMED), and 2,565 o.o.d test (WASE Normals) frames. All images have resolution 112$\times$112. Segmentation masks of the ultrasound sector were automatically generated and checked for quality manually\footnote{The automatic mask generation was implemented by an in-house proprietary software that uses classical image processing techniques.}.

\begin{table}[t!]
\scriptsize
\centering
\caption{Comparison of Augmentation methods on TMED and WASE Normals dataset. \textbf{Bold} indicates best performance} 
\begin{tabular}{lC{14mm}C{11mm}C{11mm}C{11mm}C{14mm}C{11mm}C{11mm}C{11mm}} \toprule
 & \multicolumn{4}{c}{TMED (i.d)} & \multicolumn{4}{c}{WASE Normals (o.o.d)} \\
 & Accuracy & F1 & \%E & \%F & Accuracy & F1 & \%E & \%F \\ 
 \cmidrule(lr){0-0}
 \cmidrule(lr){2-5}
 \cmidrule(lr){6-9}
Baseline & 97.7 & 97.5 & 77.9 & 92.3 & 88.7 & 88.0 & 79.5 & 94.3 \\
Black & 96.2 & 95.7 & 81.8 & 97.4 & 89.8 & 89.4 & 80.8 & 96.8 \\
Noise & 95.5 & 95.0 & 83.7 & 97.4 & 89.3 & 88.8 & 81.9 & 96.5 \\
Shuffle & 96.6 & 96.2 & 82.6 & 96.8 & 89.9 & 89.4  & 82.1 & 96.2 \\
Bokeh~\cite{li2022semantic} & 97.2 & 96.9 & 77.6 & 93.9 & 87.9 & 87.1 & 78.8 & 95.4 \\
CutMix \cite{yun2019cutmix} & \textbf{97.9} & \textbf{97.7} & 70.3 & 87.6 & 89.1 & 88.6 & 73.2 & 90.9 \\
SMA \cite{kwon2024learning} & 97.2 & 96.9 & 82.8 & 95.3 & 88.0 & 86.8 & 81.9 & 96.5 \\ 
BackMix & 96.9 & 96.2 & \textbf{86.2} & \textbf{97.8} & \textbf{92.4} & \textbf{92.1} & \textbf{85.6} & \textbf{97.8} \\ \bottomrule
 &       &       &         &        &    &    \\ [-1ex] 
\end{tabular}
\label{aug_quant}
\end{table}

\subsection{Implementation \& Training}

The baseline ResNet18 and training were implemented in PyTorch and BackMix in Numpy. All models were trained for 100 epochs on a NVIDIA GeForce RTX 2080 Ti with Adam optimiser, batch size of 64, and learning rate of 1e-3. Weights from the epoch with the highest validation accuracy were saved. For reliable performance estimates, all models were trained 3 times with 3 random seeds (shared across experiments), the mean scores were used for quantitative analysis and the model with median performance was used for qualitative analysis. Hyperparameters were tuned on a held-out validation set and set for all experiments. Alongside the attention-based metrics, we also report mean accuracy, precision, recall and F1-score to evaluate classification performance. 

\begin{figure}[t!]
    \centering
    \includegraphics[width=1\linewidth]{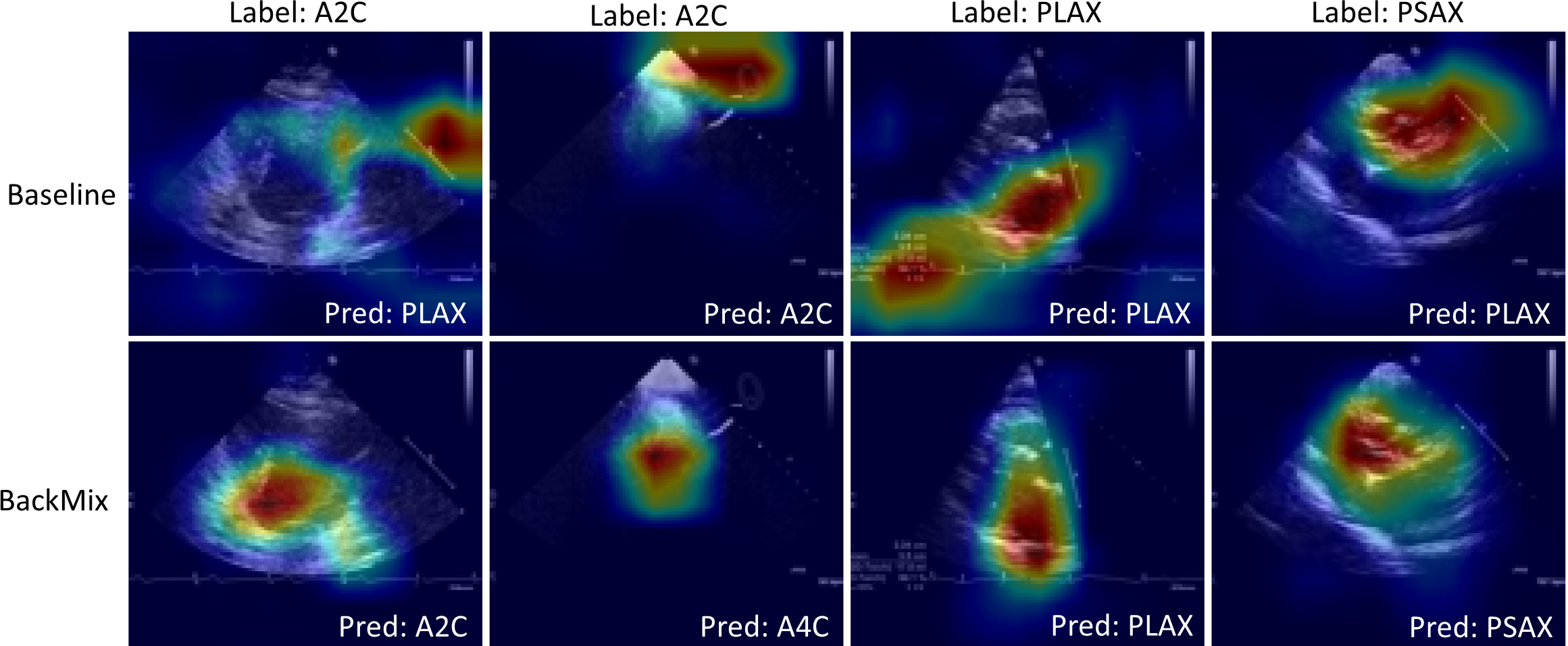}
    \caption{Qualitative results on TMED with GradCAM heatmaps.}
    \label{fig:tmed_qual}
\end{figure}

\begin{table}[b!]
\scriptsize
\centering
\caption{Semi-supervised classification for different amounts of supervision on an out-of-distribution dataset (Wase Normals).}
\begin{tabular}{L{21mm}L{9mm}C{7mm}C{11mm}C{11mm}C{11mm}C{11mm}C{11mm}C{11mm}} \toprule
 & $f$ & $\lambda$ & \multicolumn{1}{c}{Accuracy} & \multicolumn{1}{c}{Precision} & \multicolumn{1}{c}{Recall} & \multicolumn{1}{c}{F1} & \multicolumn{1}{c}{\%E} & \multicolumn{1}{c}{\%F} \\ 
\toprule
Baseline & 0 & - & 88.7 & 91.2 & 87.1 & 88.0 & 79.5 & 94.3 \\ \toprule
\multirow{6}{*}{+ BackMix} & 1 & 0 & 92.4 & 92.9 & 91.7 & 92.1 & 85.6 & 97.8 \\
 & 0.5 & 0 & 91.7 & 92.9 & 90.6 & 91.4 & 84.9 & 98.0 \\
 & 0.2 & 0 & 91.2 & 92.3 & 90.1 & 90.7 & 84.4 & 97.4 \\
 & 0.1 & 0 & 90.4 & 92.2 & 89.5 & 90.2 & 84.2 & 97.0 \\
 & 0.05 & 0 & 90.8 & 92.0 & 89.5 & 90.2 & 83.0 & 97.4 \\
 & 0.01 & 0 & 90.3 & 91.7 & 89.2 & 89.6 & 81.4 & 96.4 \\ \toprule
\multirow{2}{*}{+ wBackMix} & 0.05 & 1 & 91.4 & 92.4 & 90.5 & 91.1 & 83.3 & 97.2 \\
 & 0.05 & 2 & 91.3 & 92.5 & 90.2 & 90.9 & 83.7 & 97.5 \\ \bottomrule
 &       &       &         &        &    &    \\ [-1ex] 
\end{tabular}
 \label{wase_quant}
\end{table}

\subsection{Comparison to Existing Methods and Ablation Study}
\begin{figure}[t!]
    \centering
    \includegraphics[width=1\linewidth]{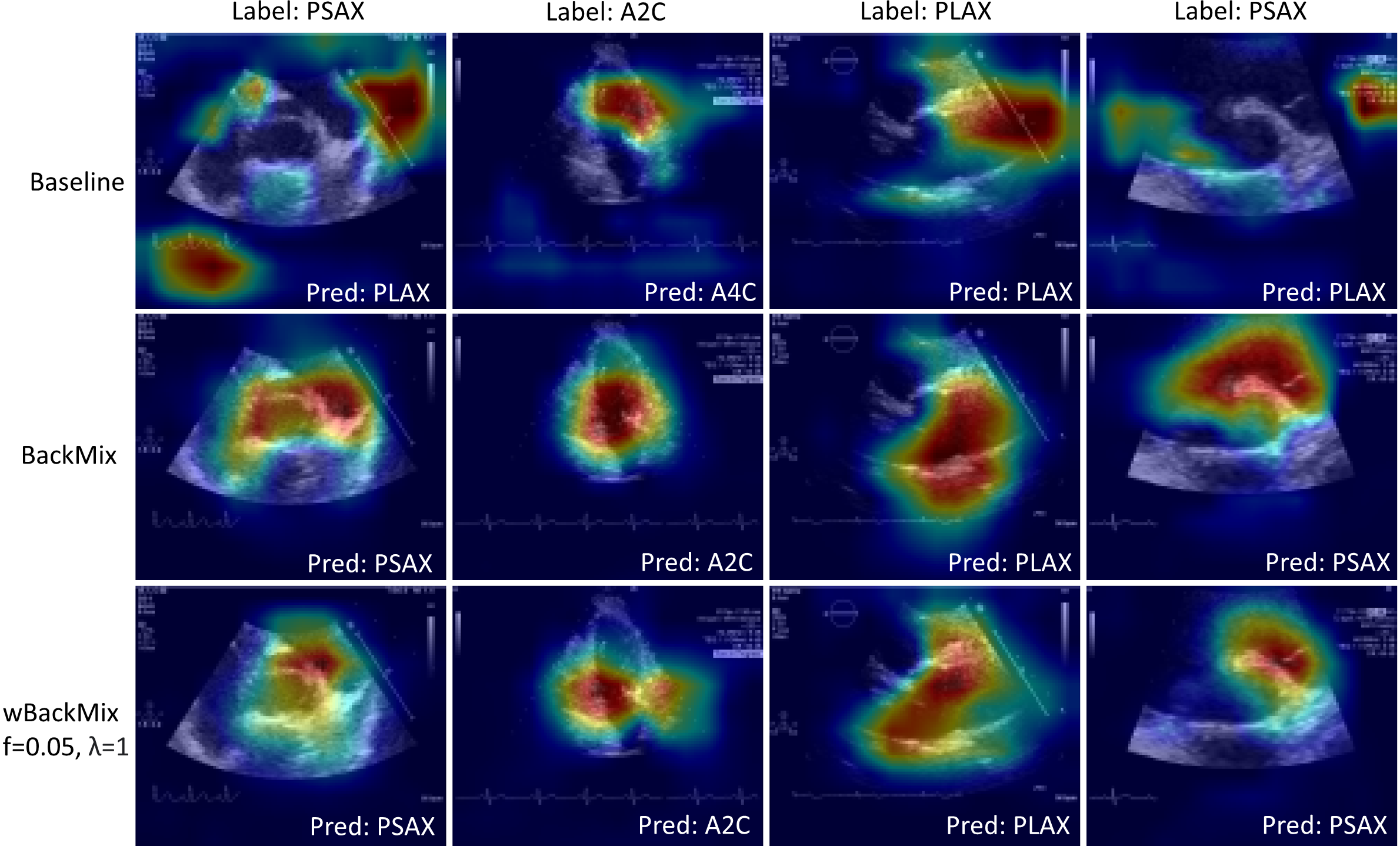}
    \caption{Qualitative results on WASE with GradCAM heatmaps.}
    \label{fig:wase_qual}
\end{figure}

We validated BackMix by comparing to various background-based configurations and augmentations in literature. These are: (1) `Black', where the background is filled with zero values; (2) `Noise', where the background is filled with random noise sampled from a uniform distribution; (3) `Shuffle', where the background pixels are randomly arranged; (4) Bokeh~\cite{li2022semantic}, a method using background blur; (5) CutMix~\cite{yun2019cutmix}, an augmentation strategy where pairs of images are mixed with a soft label; (6) SMA~\cite{kwon2024learning}, a contrastive learning method which separates object and background in feature space without segmentation masks. Quantitative and qualitative results are presented in Table~\ref{aug_quant} and Figure~\ref{fig:tmed_qual}, respectively. We focus on a single architecture (ResNet18) because the data used in this work is typically paired with that model~\cite{huang2022tmed,wessler2023automated}. 
\\
\indent The decreased classification performance on TMED for background augmentation methods is expected as the shortcuts aiding performance are not learnt due to improved sector attention (higher \%E and \%F). Augmentation methods which significantly alter images (`Black', `Noise') fare worst in both i.d and o.o.d test sets due to the incurred distribution shift. BackMix reduces this distribution shift as all backgrounds contain similar patterns and pixel intensities. In comparison to other methods, BackMix attends the sector best (high \%E and \%F), enabling learning generalisable representations of the heart. This is reflected in the highest classification performance on the o.o.d dataset. 
\\
\indent In Table~\ref{wase_quant}, we validate BackMix and wBackMix in semi-supervised classification under different amounts of supervision on the o.o.d dataset, and present a variety of examples in Figure~\ref{fig:wase_qual}. As the proportion of data to which BackMix is applied decreases, the performance naturally decreases. High performance is maintained when using 5-10\% supervision with an accuracy under 91\% and an F1 score over 90\%. Performance increases when wBackMix is applied at 5\% supervision, achieving an accuracy and F1 score comparable to BackMix at 20\% supervision. The weighting value used appears to not have a significant impact, and we would recommend a grid search to identify the best configuration. \\
\indent To further assess our semi-supervised approach, we explore in an ablation study to what extent the random selection of supervised training samples impacts accuracy and focus. We run 5 BackMix experiments (f=0.05) with a fixed random seed, but different non-overlapping supervised samples. We find standard deviations of $\pm$0.55 for accuracy, $\pm$0.55 for F1, $\pm$1.01 for \%E, and $\pm$0.70 for \%F suggesting that the choice of samples has minimal impact on performance. 

\section{Conclusion}
Echocardiograms contain imaging data in a sector, and non-imaging features that may induce shortcut learning outside the sector. Typically, image analysis methods first need to pre-process and remove background features by applying masks. However this is prone to errors, requires labels and is often computationally expensive. In this paper, we propose BackMix and wBackMix, two augmentation methods which encourage any classification network to focus on imaging data, without the need for a mask during inference. Our results demonstrate that networks trained with BackMix are able to focus more on the sector and ignore spurious correlations in the background, even when augmentation is applied to as few as 5\% of training examples. We aim to extend BackMix in the future by performing augmentation in feature space, without needing sector masks. 

\begin{credits}
\subsubsection{\discintname}
The authors have no competing interests to declare.
\end{credits}

\bibliographystyle{splncs04}
\bibliography{Paper-2248.bib}

\end{document}